\documentclass{article}

\usepackage{microtype}
\usepackage{graphicx}
\usepackage{booktabs} % for professional tables

\usepackage{hyperref}

\usepackage[accepted]{icml2025}

\usepackage{amsmath}
\usepackage{amssymb}
\usepackage{mathtools}
\usepackage{amsthm}

\usepackage[capitalize,noabbrev]{cleveref}

\theoremstyle{plain}

\theoremstyle{definition}

\theoremstyle{remark}

\usepackage[textsize=tiny]{todonotes}

\icmltitlerunning{Submission and Formatting Instructions for ICML 2025}

\usepackage{tikz}
\usepackage{pgfplots}  % Pour les graphiques avec TikZ

\pgfplotsset{compat=1.18}

\usepackage{subcaption}
\usepackage{multirow}

\definecolor{lila}{RGB}{128,0,128}
\definecolor{bluesky}{RGB}{0,149,255}
\definecolor{olive}{rgb}{0.17,0.59,0.20}
\definecolor{green}{rgb}{0.2,0.6,0.2}
\definecolor{blue}{rgb}{0.2,0.6,0.8}

\icmltitlerunning{Resampling Augmentation for Time Series Contrastive Learning}

\usepackage{tikz}
\usepackage{pgfplots}
\usepackage{amsmath,amssymb,amsfonts}
\usepackage{xcolor}
\usepackage[export]{adjustbox}
\pgfplotsset{compat=1.18}

\usetikzlibrary{calc}
\usetikzlibrary{positioning}
\usetikzlibrary{shapes.geometric}
\usetikzlibrary{fit, backgrounds}

\definecolor{my_blue}{RGB}{31, 65, 89}
\definecolor{my_green}{HTML}{99db00}
\definecolor{my_red}{HTML}{ff611a}

\newcommand{\figscale}{1.0} % global scale factor for all TikZ figures

\def\FontSizeNormal{\small}
\def\FontSizeSmall{\scriptsize}
\def\FontSizeTiny{\tiny}
\def\encwidth{1cm}  % width of trapezium (1.6cm * 0.4)
\def\encwideh{0.5cm} % wide height of trapezium (left) (0.9cm * 0.4)
\def\encnarrowh{0.25cm}  % narrow height of trapezium (right) (0.5cm * 0.4)

\def\plotOrigX{-1cm} % -8cm * 0.4
\def\plotOrigY{0cm}
\def\plotSubX{3.6cm}  % -1cm * 0.4
\def\plotSubY{0cm}
\def\plotOneX{7cm}   % 7cm * 0.4
\def\plotOneY{1cm}    % 2cm * 0.4
\def\plotTwoX{7cm}   % 7cm * 0.4
\def\plotTwoY{-1cm}   % -2cm * 0.4

\def\axisWidth{1.8cm}  % 4cm * 0.4
\def\axisHeight{0.8cm} % 2cm * 0.4
\def\plotMarkSize{0.6pt} % 1.5pt * 0.4 (New)
\def\axisTitleYShift{-0.1cm} % yshift for axis titles (New)
\def\tsBgShiftLarge{0.42cm}     % 0.35cm * 0.4
\def\tsBgShiftMedLarge{0.36cm}  % 0.30cm * 0.4
\def\tsBgShiftMedium{0.30cm}    % 0.25cm * 0.4
\def\tsBgShiftSmall{0.24cm}     % 0.20cm * 0.4
\def\tsBgShiftTiny{0.18cm}      % 0.15cm * 0.4
\def\tsBgShiftMicro{0.12cm}     % 0.10cm * 0.4
\def\tsBgShiftNano{0.06cm}      % 0.05cm * 0.4

\def\resampleNodeOffsetX{1.7cm}    % 4.25cm * 0.4
\def\resampleNodeMinSize{0.32cm}   % 0.8cm * 0.4

\def\projectorMinWidth{0.24cm}    % 0.6cm * 0.4
\def\projectorMinHeight{1.0cm}   % 2.5cm * 0.4
\def\aggregatorMinWidth{0.32cm}   % 0.8cm * 0.4
\def\componentInnerSep{0.1cm} % 5pt * 0.4 (New)
\def\componentStyleInnerSep{4pt}  % 10pt * 0.4 (New)
\def\simclrScopeShiftX{0.4cm}   % 1.5cm * 0.4 Shift for the entire SimCLR framework scope

\def\encOneY{1cm}     % 2cm * 0.4
\def\encTwoY{-1cm}    % -2cm * 0.4 % Symmetrical to encOneY
\def\encLabelTextWidth{0.72cm} % 1.8cm * 0.4 (New)
\def\encLabelOffsetY{-0.37cm}  % -12pt * 0.4 % Offset for "ResNet Encoder" label
\def\hSetLabelOffsetX{0.12cm}  % 0.3cm * 0.4 % Offset for h_set_label (e.g., h^1)
\def\aggOffsetX{0.12cm}        % 0.3cm * 0.4 % Offset for aggregator node
\def\HLabelOffsetX{0.2cm}     % 0.5cm * 0.4 % Offset for H_label_node (e.g., H_1)
\def\projOffsetX{0.2cm}       % 0.5cm * 0.4 % Offset for projector node (MLP Proj.)
\def\ZLabelOffsetX{0.2cm}     % 0.5cm * 0.4 % Offset for Z_label (e.g., Z_1)
\def\ZLabelFitExtend{0.2cm}  % 0.5cm * 0.4 % Extension for fitting background around Z labels

\def\arrowToResampleNodeOffsetX{0.12cm} % 0.3cm * 0.4
\def\arrowToEncoderOffsetX{0.12cm}    % 0.3cm * 0.4

\def\contrastiveLossOffsetX{0.38cm}    % 1.2cm * 0.4 % Offset for contrastive_loss node
\def\legendNontrainableOffsetY{-1cm} % -0.2cm * 0.4
\def\legendTrainableSpacing{0.04cm}    % 0.1cm * 0.4
\def\viewGenBgInnerSep{14pt}            % 15pt * 0.4
\def\viewGenBgMinHeight{4.5cm}         % 3.5cm * 0.4
\def\simclrBgInnerSep{4.4pt}           % 11pt * 0.4
\def\simclrBgMinHeight{4.5cm}          % 3.5cm * 0.4 % Same as viewGenBgMinHeight for consistency
\def\tsBackgroundInnerSep{0.15cm}       % 6pt * 0.4

\tikzset{
  encoderPic/.pic = {
    \coordinate (-left) at (0,0);
    \draw [fill=orange!20] (0,-\encwideh) -- (0,\encwideh) -- (\encwidth,\encnarrowh) -- (\encwidth,-\encnarrowh) -- cycle ;
    \coordinate (-center) at (\encwidth/2,0);
    \coordinate (-right) at (\encwidth,0);
    \coordinate (-top) at (\encwidth/2,\encwideh);
    \coordinate (-bottom) at (\encwidth/2,-\encwideh);
  },
  encoderPicEMA/.pic = {
    \coordinate (-left) at (0,0);
    \draw [draw=black, fill=gray!20, dotted] (0,-\encwideh) -- (0,\encwideh) -- (\encwidth,\encnarrowh) -- (\encwidth,-\encnarrowh) -- cycle;
    \coordinate (-center) at (\encwidth/2,0);
    \coordinate (-right) at (\encwidth,0);
    \coordinate (-top) at (\encwidth/2,\encwideh);
    \coordinate (-bottom) at (\encwidth/2,-\encwideh);
  },
  tsBackground/.style = {draw=black, fill=white, rounded corners, inner sep=\tsBackgroundInnerSep, thin} % Outline style for time series background
}

\begin{document}

\twocolumn[
\icmltitle{Resampling Augmentation for Time Series Contrastive Learning: \\
           Application to Remote Sensing}



\icmlsetsymbol{equal}{*}

\begin{icmlauthorlist}
\icmlauthor{Antoine Saget}{icube}
\icmlauthor{Baptiste Lafabregue}{icube}
\icmlauthor{Antoine Cornuéjols}{agroparistech}
\icmlauthor{Pierre Gançarski}{icube}
\end{icmlauthorlist}

\icmlaffiliation{icube}{ICube, University of Strasbourg, Strasbourg, France}
\icmlaffiliation{agroparistech}{AgroParisTech, Paris, France}

\icmlcorrespondingauthor{Antoine Saget}{antoinesaget19@gmail.com}

\icmlkeywords{contrastive learning, time series, remote sensing, data augmentation, cropland classification, self-supervised learning}

\vskip 0.3in
]



\printAffiliationsAndNotice{}  % leave blank if no need to mention equal contribution

\begin{abstract}
    Given the abundance of unlabeled Satellite Image Time Series (SITS) and the scarcity of labeled data, contrastive self-supervised pretraining emerges as a natural tool to leverage this vast quantity of unlabeled data. However, designing effective data augmentations for contrastive learning remains challenging for time series. We introduce a novel resampling-based augmentation strategy that generates positive pairs by upsampling time series and extracting disjoint subsequences while preserving temporal coverage. We validate our approach on multiple agricultural classification benchmarks using Sentinel-2 imagery, showing that it outperforms common alternatives such as jittering, resizing, and masking. Further, we achieve state-of-the-art performance on the S2-Agri100 dataset without employing spatial information or temporal encodings, surpassing more complex masked-based SSL frameworks. Our method offers a simple, yet effective, contrastive learning augmentation for remote sensing time series.
\end{abstract}

\section{Introduction}
\label{sec:intro}

Every five days, the Sentinel-2 satellite constellation \cite{drusch2012sentinel,gascon2017copernicus} captures multispectral images of Earth's entire surface at 10-meter resolution, generating an unprecedented amount of data of our planet's changing landscapes. However, a major challenge lies in the scarcity of labeled data. While large volumes of raw Satellite Images Time Series (SITS) data are available, labeling them is costly, time-consuming, expert-dependent, and often domain-specific. Consequently, only a fraction of this data can be leveraged in fully supervised frameworks, leaving the rest unused.

Self-Supervised Learning (SSL) methods offer a promising solution to exploit these large, unlabeled SITS datasets. By learning meaningful representations from unlabeled data, SSL can achieve higher accuracy on downstream tasks with fewer labeled samples \cite{henaff2020data}. Generally, SSL methods can be divided into two main categories: \textit{contrastive methods}, which rely on bringing closer together pairs of similar samples (generally created through data augmentation) in representation space, and \textit{generative} or mask-based methods, which focus on reconstructing missing (artificially masked) information in the data. 

Masking strategies have been extensively studied for time series data. Although the performance achieved is interesting, these methods can be difficult to implement for spatio-temporal data. They have a high computational cost, and the definition of tokens for transformer-based models is highly dependent on the dataset used. In addition, masking strategies have better performance when combined with contrastive methods \cite{cheng2023timemae}. Hence, further study of contrastive methods is needed. However, contrastive learning methods for SITS remain under-explored due to the difficulty of designing robust augmentations for time series data. Indeed, a key component of contrastive methods is the formation of positive sample pairs: two views of the same underlying instance that should be mapped close together in the representation space. In computer vision, standard data augmentations such as cropping, rotation, and color jittering have been well studied \cite{chen2020simple, chen2020big}. Yet for time series data, including SITS, designing similarly effective augmentations is less straightforward and remains an active research topic~\cite{liu2024guidelines}. \\
In this paper, we make three main contributions: 

First, we introduce a novel, resampling-based, augmentation technique for time series contrastive learning. This straightforward approach generates two views of a time series by upsampling the original sequence, then extracting two disjoint subsequences from it while maintaining temporal coverage across the full temporal range.

Second, we experimentally show that our resampling augmentation outperforms traditional time series augmentations (namely jittering, resizing, and masking) for contrastive learning on satellite image time series data. Furthermore, despite its simplicity and without relying on spatial information or temporal positional encodings, our approach achieves state-of-the-art performance on the S2-Agri100 dataset \cite{garnot2020satellite, yuan2022sits} for satellite image time series classification of agricultural fields, surpassing methods based on masked input reconstruction.

Third, we investigate the impact of pretraining data distribution in SITS. We show that pretraining on unlabeled data from the target domain (S2-Agri100 dataset) rather than a different domain (SITS-Former dataset) enables a simple logistic regression to outperform state-of-the-art models trained on the SITS-Former dataset. Also, we observe a minimal performance difference between full finetuning and linear evaluation, suggesting that feature quality plays a greater role than classifier complexity, and that collecting large quantities of unlabeled data from the target domain can be as valuable as obtaining small quantities of labels.

The remainder of this paper is organized as follows: \cref{sec:related_works} reviews related work in contrastive learning, self-supervised learning for remote sensing, and time series augmentations. \cref{sec:method} presents our resampling augmentation technique in detail. \cref{sec:setup} describes our experimental setup, detailing the datasets used for pretraining and downstream evaluation, model architectures, and training protocols. \cref{sec:results} presents our experimental results through four main analyses: (1) a comparison of different contrastive learning frameworks, (2) a comprehensive evaluation of label efficiency across multiple datasets, (3) performance benchmarking on the S2-Agri100 dataset, and (4) a small investigation of the impact of pretraining data distribution. Finally, \cref{sec:conclusion} concludes with a discussion of limitations and future research directions.

All code for models, training, evaluation, and datasets preprocessing is available\footnote{\url{https://github.com/antoinesaget/ts_ssl} and \url{https://github.com/antoinesaget/sits_dl_preprocess}}.

\section{Related Works}
\label{sec:related_works}

\subsection{Contrastive Learning Frameworks}

The core idea of Contrastive Self-Supervised Learning is to bring the representations of similar samples (positive pairs) closer together in the embedding space. These positive pairs are typically created by applying different data augmentations to the same input sample. However, focusing solely on making positive pairs similar can lead to representation collapse, where the model maps all inputs to the same representation. Different frameworks address this challenge in distinct ways. SimCLR \cite{chen2020simple} uses in-batch negatives, treating other samples within the batch as negative examples and pushing them apart in representation space. MoCo \cite{he2019momentum} extends this approach with a memory bank to include more negative examples and using a momentum-updated encoder to generate consistent representations. However, these methods can suffer from false negatives when samples from the same class are mistakenly pushed apart. BYOL \cite{grill2020bootstrap} avoids negative examples entirely and prevents collapse using a momentum encoder and asymmetric branches (prediction head in one branch, none in the other). SimSiam \cite{chen2021exploring} simplifies BYOL by showing that the momentum encoder is not necessary. VICReg \cite{bardes2021vicreg} directly prevents collapse through variance and covariance regularization terms in the loss. Other approaches include SwAV \cite{caron2020unsupervised}, which uses online clustering, and Barlow Twins \cite{zbontar2021barlow}, which maximizes the independence of features.

\subsection{Self-Supervised Learning in Remote Sensing}

Recent advances in Self-Supervised Learning (SSL) for remote sensing have largely focused on developing Remote Sensing Foundation Models (RSFMs). These models all contribute towards the ideal of universal representations applicable across any satellite sensors, spatial scales, geographical locations, temporal resolutions, and downstream tasks. Two main approaches have emerged: masked modeling and contrastive learning.

\textit{Masked modeling approaches}, inspired by the success of masked autoencoders (MAE) \cite{he2022masked} in computer vision, have been widely adopted. SatMAE \cite{cong2022satmae} and Prithvi \cite{jakubik2023foundation} use temporal and spectral encodings alongside traditional spatial encodings to handle the multi-modal nature of satellite data. ScaleMAE \cite{reed2022scale} contributes towards scale invariance by separately reconstructing low and high-frequency components of masked regions.

These spatio-temporal MAE approaches face an inherent computational challenge: the cubic growth in the number of tokens (width × height × time) added to the quadratic growth of self-attention in transformer models with respect to the number of tokens. This requires a trade-off, and most models prioritize spatial coverage at the expense of temporal depth (e.g., SatMAE \cite{cong2022satmae} is limited to 3 timesteps). Presto \cite{tseng2023lightweight} takes the opposite approach by focusing exclusively on the temporal dimension without spatial context, enabling it to process much longer time series. Trained on a large-scale worldwide dataset of 20M time series combining Sentinel-1 SAR, Sentinel-2 multispectral, ERA meteorological data, and more, it treats each pixel independently and applies temporal and spectral masking on sequences of 12 timesteps covering 12 months, demonstrating competitive performance even against methods that leverage spatial information.

\textit{Contrastive learning} offers an alternative approach. Seasonal Contrast (SeCo) \cite{manas2021seasonal} trains on large-scale Sentinel-2 imagery using three simultaneous objectives with separate projection heads from a shared embedding space: one head learns invariance to standard image augmentations (random cropping, color jittering, flipping), another learns invariance to seasonal changes by bringing closer images of the same location at different times, and a third combines both types of invariance. While this results in time-aware representations, the model cannot directly process time series as input. SSL4EO-S12 \cite{wang2022ssl4eo} extends this work to multi-modal data (Sentinel-1 and Sentinel-2) while evaluating various contrastive frameworks (MoCo, DINO, MAE).

Recent work has focused on improving the universality of these models. DOFA \cite{wang2023decur} generates dynamic weights to adapt to unseen sensors, trained on a diverse dataset spanning Sentinel-2 multispectral, Sentinel-1 SAR, EnMAP hyperspectral, and high-resolution aerial imagery. SkySense \cite{guo2024skysense} employs multi-granularity contrastive learning to create embeddings effective at pixel, object, and image scales. Its training data combines high-resolution WorldView-3/4 imagery with temporal sequences from Sentinel-1/2. Despite being one of the largest RSFM to date (25,000 NVIDIA A100 GPU hours) with the longest time series support, SkySense notably does not incorporate temporal augmentations during contrastive learning.

\subsection{Time Series Augmentations}

While masked modeling approaches do not require data augmentations, contrastive methods traditionally rely on spatial augmentations like cropping, rotation, and color jittering that are not directly applicable to time series data. The diversity of time series data - from satellite observations to electrocardiograms and stock prices - makes designing universal augmentations particularly challenging due to their varying characteristics, sampling frequencies, and lengths.

\citeauthor{liu2024guidelines} \yrcite{liu2024guidelines} provide a comprehensive analysis of time series augmentations for contrastive learning, evaluating eight common transformations: jittering (adding random noise), scaling (multiplying by a random factor), flipping (reversing values), permutation (shuffling segments), resizing (temporal interpolation), time masking (zeroing random timesteps), frequency masking (filtering frequency components), and time neighboring (selecting adjacent windows). They identify which augmentations are most effective for different types of time series based on properties such as seasonality, trend, and noise levels. 

\section{Resampling augmentation}
\label{sec:method}

% Define all coordinates at the beginning for easy modification
\pgfplotstableread{
x   y_orig
0   5.00
1   15.00
2   32.50
3   28.50
4   12.50
5   7.50
6   11.00
7   2.50
}\originaldata

\pgfplotstableread{
x   y_up
0   5
1   10
2   15  
3  23.75
4   32.5
5   30.5
6   28.5
7   20.5
8   12.5
9   10
10  7.5
11  9.25
12  11
13  6.75
14  2.5
 }\upsampleddata
 
\pgfplotstableread{
x   y
0   5
3   23.75
6   28.5
9   10
14  2.5 
}\subsampleone

\pgfplotstableread{
x   y
1   10
4   32.5
7   20.5
10  7.5
12  11
13  6.75
}\subsampletwo

\pgfplotstableread{
x   y
0   5
1   12.5
2   28.5   
3   28
4   14.4
5   8.8
6   5.5
7   2.5
}\finalviewone
     
\pgfplotstableread{
x   y
0   10
1   16
2   30.3
3   25
4   15
5   9.3
6   9.7
7   6.5
}\finalviewtwo

% Common axis style
\pgfplotsset{
    common/.style={
        width=10cm,
        height=3.5cm,
        axis lines=middle,
        xmin=-0.1,
        xmax=7.3,
        ymin=0,
        ymax=35,
        xtick={0,1,2,3,4,5,6,7},
        ytick=\empty,
        enlarge x limits=false,
        axis x line=middle,
        axis y line=middle,
        axis y line*=left,
        every axis x label/.append style={at={(ticklabel* cs:1)}, anchor=north},
        tick label style={font=\footnotesize},
        xlabel style={font=\footnotesize},
        ylabel style={font=\footnotesize},
        legend style={at={(0.95,0.95)}, anchor=north east, font=\scriptsize},
        legend cell align={left},
        title style={font=\footnotesize, align=center}
    }
}

\begin{figure}[t!b]
    \centering
    \begin{subfigure}{\columnwidth}
        \begin{tikzpicture}
            \begin{axis}[
                width=9cm, % largeur du graphique
                height=3.7cm, % hauteur du graphique
                axis lines=middle, % lignes des axes passant par le milieu
                xlabel={$T$}, % label de l'axe des x
               % ylabel={$V$}, % label de l'axe des y
                xmin=-0.1, xmax=7.7, % limites de l'axe des x
                ymin=0, ymax=35, % limites de l'axe des y
                xtick={0,1,2,3,4,5,6,7}, % positions des graduations sur l'axe des x
                ytick=\empty,  % positions des graduations sur l'axe des y
               % grid=major, % ajout d'une grille
               % grid style=dashed, % style de la grille
                enlarge x limits=false, % pas d'élargissement automatique des limites
                axis x line=middle, % axe des x au centre
                axis y line=middle, % axe des y au centre
                axis y line* = left, % décaler l'axe des ordonnées vers la gauche
                every axis x label/.append style={at={(ticklabel* cs:1)}, anchor=north}, % aligner le label de l'axe x
                tick label style={font=\footnotesize}, % réduire la taille des graduations
                xlabel style={font=\footnotesize}, % réduire la taille du label de l'axe x
                ylabel style={font=\footnotesize}, % réduire la taille du label de l'axe y
                legend style={at={(1, 1)}, anchor=north east, font=\scriptsize}, % position et taille de la légende
                legend cell align={left}, % alignement des entrées de la légende
             %   title={Original data}, % titre du graphique
                title style={font=\footnotesize, align=center}
            ]
                % Tracé des points reliés par une ligne en rouge avec une légende
                  \addplot[color=black, mark=*] table {\originaldata};
           %     \addplot[color=black, mark=*] 
        %     coordinates { (0,5)  (1,15)  (2,32.5)   (3,28.5) (4,12.5) (5,7.5) (6,11) (7,2.5)};    
                \addlegendentry{S} % Ajout de la légende
            \end{axis}
        \end{tikzpicture}
        \caption{Original time serie} 
        \label{fig:Original}
    \end{subfigure}

    \vspace{0.3cm}
    
    \begin{subfigure}{\columnwidth}
        \begin{tikzpicture}
            \begin{axis}[
                width=9cm, % largeur du graphique
                height=3.7cm, % hauteur du graphique
                axis lines=middle, % lignes des axes passant par le milieu
                xlabel={$T$}, % label de l'axe des x
               % ylabel={$V$}, % label de l'axe des y
                xmin=-0.1, xmax=15, % limites de l'axe des x
                ymin=0, ymax=35, % limites de l'axe des y
                xtick={0,1,2,3,4,5,6,7,8,9,10,11,12,13,14}, % 14 graduations sur l'axe des x
                ytick=\empty,   %      ytick={0,10,20,30}, % positions des graduations sur l'axe des y
             %   grid=major, % ajout d'une grille
              %  grid style=dashed, % style de la grille
                enlarge x limits=false, % pas d'élargissement automatique des limites
                axis x line=middle, % axe des x au centre
                axis y line=middle, % axe des y au centre
                axis y line* = left, % décaler l'axe des ordonnées vers la gauche
                every axis x label/.append style={at={(ticklabel* cs:1)}, anchor=north}, % aligner le label de l'axe x
                tick label style={font=\footnotesize}, % réduire la taille des graduations
                xlabel style={font=\footnotesize}, % réduire la taille du label de l'axe x
                ylabel style={font=\footnotesize}, % réduire la taille du label de l'axe y
                legend style={at={(1, 1)}, anchor=north east, font=\scriptsize}, % position et taille de la légende
                legend cell align={left}, % alignement des entrées de la légende
             %   title={1 - Upsampling}, % from N\_timesteps to 2xN\_timesteps}, % titre du graphique
                title style={font=\footnotesize, align=center} % style du titre
            ]
                \addplot[color=blue, mark=*] table {\upsampleddata};
              %  coordinates { (0,5)  (1,10) (2,15)  (3,23.75) (4,32.5) (5,30.5)  (6,28.5)  (7,20.5) (8,12.5) (9,10)  (10,7.5)  (11,9.25) (12,11)  (13,6.75) (14,2.5) };       
                 \addlegendentry{$S_{up}$}
            \end{axis}
        \end{tikzpicture}
        \caption{Upsampled time serie}
        \label{fig:UP}
    \end{subfigure}

    \vspace{0.3cm}
    
    \begin{subfigure}{\columnwidth}
        \begin{tikzpicture}
            \begin{axis}[
           width=9cm, % largeur du graphique
                height=3.7cm, % hauteur du graphique
                axis lines=middle, % lignes des axes passant par le milieu
               xlabel={$T$}, % label de l'axe des x
               % ylabel={$V$}, % label de l'axe des y
                xmin=-0.1, xmax=15, % limites de l'axe des x
                ymin=0, ymax=35, % limites de l'axe des y
                xtick={0,1,2,3,4,5,6,7,8,9,10,11,12,13,14}, % 14 graduations sur l'axe des x
                ytick=\empty, % Supprimer les graduations sur l'axe vertical
             %   grid=major, % ajout d'une grille
          %      grid style=dashed, % style de la grille
                enlarge x limits=false, % pas d'élargissement automatique des limites
                axis x line=middle, % axe des x au centre
                axis y line=middle, % axe des y au centre
                axis y line* = left, % décaler l'axe des ordonnées vers la gauche
                every axis x label/.append style={at={(ticklabel* cs:1)}, anchor=north}, % aligner le label de l'axe x
                tick label style={font=\footnotesize}, % réduire la taille des graduations
                xlabel style={font=\footnotesize}, % réduire la taille du label de l'axe x
                ylabel style={font=\footnotesize}, % réduire la taille du label de l'axe y
                legend style={at={(1, 1)}, anchor=north east, font=\scriptsize}, % position et taille de la légende
                legend cell align={left}, % alignement des entrées de la légende
             %   title={2 - Subsampling}, % titre du graphique
                title style={font=\footnotesize, align=center} % style du titre
            ]
                % Tracé des points reliés par une ligne en rouge avec une légende
                      
                 \addplot[color=white!40!blue!40!, mark=*] table {\upsampleddata};
            %     coordinates { (0,5)          (1,10)            (2,15)       (3,23.75) (4,32.5) (5,30.5)  (6,28.5)  (7,20.5) (8,12.5) (9,10)  (10,7.5)  (11,9.25) (12,11)  (13,6.75) (14,2.5) };
         \addlegendentry{$S_{up}$} % Ajout de la légende
                
                 % Tracé des points reliés par une ligne en rouge avec une légende
               \addplot[color=black!30!green, mark=*] table {\subsampleone};
          % coordinates { (0,5)      (3,23.75)    (6,28.5)   (9,10)   (14,2.5) };
        
                 \addlegendentry{$S_{sub}^1$} % Ajout de la légende
                
                % Tracé de la deuxième courbe en rouge prenant un point sur deux
                \addplot[color=white!30!red, mark=*] table {\subsampletwo};
          %  coordinates {   (1,10)  (4,32.5)  (7,20.5)  (10,7.5)   (12,11)  (13,6.75)  };      
              \addlegendentry{$S_{sub}^2$} % Ajout de la légende pour la nouvelle courbe
               \addplot[dashed, color=gray] coordinates {(3.5,0) (3.5,33)};      
              \addplot[dashed, color=gray] coordinates {(7,0) (7,33)};
              \addplot[dashed, color=gray] coordinates {(10.5,0) (10.5,33)};
               \addplot[dashed, color=gray] coordinates {(14,0) (14,33)};
            \end{axis}
        \end{tikzpicture}
        \caption{Subsampled time series}
        \label{fig:SUB}
    \end{subfigure}

    \vspace{0.3cm}
    
    \begin{subfigure}{\columnwidth}
        \begin{tikzpicture}
            \begin{axis}[
                width=9cm, % largeur du graphique
                height=3.7cm, % hauteur du graphique
                axis lines=middle, % lignes des axes passant par le milieu
                xlabel={$T$}, % label de l'axe des x
               % ylabel={$V$}, % label de l'axe des y
                xmin=-0.1, xmax=7.5, % limites de l'axe des x
                ymin=0, ymax=35, % limites de l'axe des y
                xtick={0,1,2,3,4,5,6,7}, % positions des graduations sur l'axe des x
                ytick=\empty,  % positions des graduations sur l'axe des y
               % grid=major, % ajout d'une grille
               % grid style=dashed, % style de la grille
                enlarge x limits=false, % pas d'élargissement automatique des limites
                axis x line=middle, % axe des x au centre
                axis y line=middle, % axe des y au centre
                axis y line* = left, % décaler l'axe des ordonnées vers la gauche
                every axis x label/.append style={at={(ticklabel* cs:1)}, anchor=north}, % aligner le label de l'axe x
                tick label style={font=\footnotesize}, % réduire la taille des graduations
                xlabel style={font=\footnotesize}, % réduire la taille du label de l'axe x
                ylabel style={font=\footnotesize}, % réduire la taille du label de l'axe y
                legend style={at={(1,1)}, anchor=north east, font=\scriptsize}, % position et taille de la légende
                legend cell align={left}, % alignement des entrées de la légende
             %   title={3 - Interpolation}, % titre du graphique
                title style={font=\footnotesize, align=center}
            ]
                % Tracé des points reliés par une ligne en rouge avec une légende
                \addplot[color=gray, mark=*] table {\originaldata};
          %   coordinates { (0,5)  (1,15)  (2,32.5)   (3,28.5) (4,12.5) (5,7.5) (6,11) (7,2.5)};    
                \addlegendentry{S} % Ajout de la légende
                
                       \addplot[color=black!30!green, mark=*] table {\finalviewone};
         %   coordinates { (0,5)      (1,12.5)    (2,28.5)   (3,28.5)   (4,14.4) (5,8.8) (6,5.5) (7,2.5) };
        
                 \addlegendentry{$S^1$} % Ajout de la légende
                
                % Tracé de la deuxième courbe en rouge prenant un point sur deux
                \addplot[color=white!30!red, mark=*] table {\finalviewtwo};
        %    coordinates {   (0,10) (1,16)  (2,30.3)   (3,25) (4,15) (5,9.3) (6,9.7) (7,6.5)   };       
             \addlegendentry{$S^2$} % Ajout de la légende pour la nouvelle courbe
        
            \end{axis}
        \end{tikzpicture}

        \caption{Interpolated and aligned time series}
        \label{fig:FINAL}
    \end{subfigure}
    
    \caption{Example visualization of the resampling augmentation process.}
\end{figure}
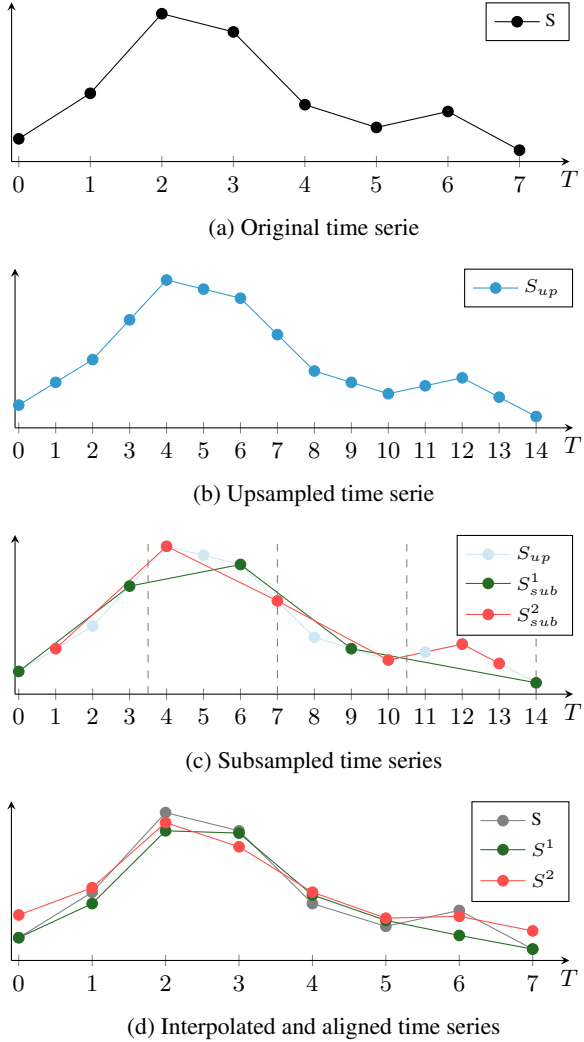 

Given an input time series $\mathbf{S} = \{s_1, \dots , s_T\} \in \mathbb{R}^{T \times C}$, where $T$ is the number of timesteps with index $\mathcal{X} = \{1, ..., T\}$ and $C$ is the number of channels, we perform the temporal resampling augmentation in three steps.

\textit{First} (Fig.~\ref{fig:UP}),  we upsample the original time series to $T_{up}$ timesteps (typically $T_{up} = 2 \times T$) using linear interpolation:
\begin{equation}
    \mathbf{S}_{up} = f_{linear}(\mathbf{S}) \in \mathbb{R}^{T_{up} \times C}
\end{equation}

\textit{Second} (Fig.~\ref{fig:SUB}), we sample two subsequences $\mathbf{S}_{sub}^1$ (resp. $\mathbf{S}_{sub}^2$) with indices $\mathcal{X}^1_{sub}$ (resp. $\mathcal{X}^2_{sub}$)  containing $T_{int}^1$ (resp. $T_{int}^2$) timesteps from $\mathbf{S}_{up}$ (typically $T_{int}^i = T / 2$). The sampling strategy follows two constraints:
\begin{itemize}
    \item The subsequences use distinct timesteps: $\mathcal{X}_{sub}^1 \cap \mathcal{X}_{sub}^2 = \emptyset$
    \item Each subsequence samples uniformly at least $\lfloor T_{int}^i/4 \rfloor$ timesteps within each quarter of $\mathbf{S}_{up}$:
\end{itemize}
\begin{equation}
    |\mathcal{X}_{sub}^i \cap \{(j-1)\frac{T_{up}}{4}, ..., j\frac{T_{up}}{4}\}| \geq \lfloor \frac{T_{int}^i}{4} \rfloor \text{, } j \in \{1,..,4\}
\end{equation}
This structured sampling constraint ensures that both sequences maintain complete temporal coverage of the original signal and prevent large temporal gaps.

\textit{Third} (see Fig.~\ref{fig:FINAL}), we resample both sequences to match the original temporal resolution:
\begin{equation}
    \mathbf{S^i} = f_{resample}(\mathbf{S_{sub}^i}) \in \mathbb{R}^{T \times C}
\end{equation}
The function $f_{resample}$ transforms a subsequence $\mathbf{S_{sub}^i}$ with timestamps $\mathcal{X}_{sub}^i$ into a time series aligned with the original timestamps $\mathcal{X}$ through two steps:
\begin{enumerate}
    \item First, we linearly rescale the timestamps $\mathcal{X}_{sub}^i$ to span the full range $[1, T]$, preserving their relative spacing. This maps the subsequence onto the same temporal range as the original series.
    \item Then, since the rescaled timestamps generally don't align with the original timestamps $\mathcal{X}$, we use linear interpolation to compute values at exactly the timestamps in $\mathcal{X}$, ensuring the resulting time series has both the same temporal resolution and temporal alignment as the input.
\end{enumerate}
This augmentation results in two distinct but similar time series that preserve the overall temporal structure, length, and alignment while introducing controlled variations, making them suitable as positive pairs for contrastive learning.

\section{Experimental Setup}
\label{sec:setup}

\subsection{Datasets}

Unlike recent works aiming to build general-purpose remote sensing foundation models \cite{tseng2023lightweight, jakubik2023foundation, reed2022scale, cong2022satmae, guo2024skysense, wang2023decur, wang2022ssl4eo, manas2021seasonal}, we focus on task-specific pretraining. While our resampling augmentation could be used in a more flexible setting with varying data shapes and temporal resolutions, in this paper we focus on demonstrating its effectiveness in a more limited setting. We pretrain each model on unlabeled data that matches the characteristics (i.e. data shape, source, location) of its target downstream task. Table \ref{tab:datasets_summary} details the pretraining and downstream datasets used in our experiments.

FranceCrops \cite{saget2024learning} is a large-scale Sentinel-2 time series dataset for agricultural parcel classification in metropolitan France. Each sample consists of 100 pixel time series sampled within a crop field's geometric bounds, with 60 temporally aligned timesteps spanning February-November 2022 across all 12 Sentinel-2 L2A bands. It is split into an unlabeled contrastive learning set ($\approx$4M samples) and labeled sets (train/val/test) containing 20 selected common crop types. A separate dataset for the Centre-Val de Loire region follows the same structure with a different subset of 20 classes, enabling evaluation of geographical generalization.

PASTIS \cite{garnot2021panoptic} is a similar agricultural parcel classification dataset but at a smaller scale with ~85k samples and 18 crop types. Unlike FranceCrops, raw PASTIS samples contain varying numbers of unaligned time series per parcel. We preprocess the dataset following the same procedure as FranceCrops to obtain aligned time series of equal length. While PASTIS also offers versions with full imagery for spatio-temporal models and Sentinel-1 data, we only use the Sentinel-2 pixel-set version in this study.

The SITS-Former pretraining dataset \cite{yuan2022sits} consists of $\approx$1.66M unlabeled Sentinel-2 time series of 24 timesteps across 10 channels sampled from California's Central Valley during 2018-2019. Each sample is a 5×5 pixel patch extracted at regular intervals from cloud-filtered ($<$10\%) Level-2A images. We process each 5x5 patch as a set of independent pixel time series, disregarding spatial relationships between pixels.

S2-Agri100 \cite{yuan2022sits} is a variant of the S2-Agri dataset \cite{garnot2020satellite} for crop type classification, sharing similar characteristics with the SITS-Former dataset (5×5 pixel patches, 24 timesteps, 10 channels, cloud-filtered Sentinel-2 bands) but located in southern France spanning January-October 2017. The dataset contains $\approx$175k test samples from a 12,100$km^2$ area, with 100 samples per class in both training and validation sets. Except for the final experiment in Section \ref{sec:pretraining_distribution}, this dataset is only used for downstream evaluation.

\begin{table*}[h!t]
\caption{Datasets characteristics for self-supervised pretraining and/or supervsied downstream task evaluation.}
\small
\begin{center}
\begin{tabular}{l||ccc||c||cc}
\multicolumn{1}{c||}{} & \multicolumn{3}{c||}{\textbf{Sample shape}} & \textbf{Pretraining} & \multicolumn{2}{c}{\textbf{Downstream task evaluation}} \\
\cline{2-7}
\textbf{Dataset} & \textbf{Nb time series} & \textbf{Nb Timesteps} & \textbf{Nb Channels} & \textbf{Nb samples} & \textbf{Nb Classes} & \textbf{Nb samples/class} \\
\hline\hline
FranceCrops                         & 100 & 60   & 12 & $\sim$5.8M   & 20 & 5--100 \\
FranceCrops CVdL                    & 100 & 60   & 12 & --           & 20 & 5--100 \\
PASTIS                              & 100 & 60   & 10 & $\sim$85k    & 18 & 5--100 \\
SITS-Former                         & 25 & 24 & 10 & $\sim$1.6M & \multicolumn{2}{c}{--}    \\
S2-Agri100                          & 25 & 24 & 10 & $\sim$120k   & 15 & 100    \\
\end{tabular}
\end{center}
\label{tab:datasets_summary}
\end{table*}

\subsection{Architectures}

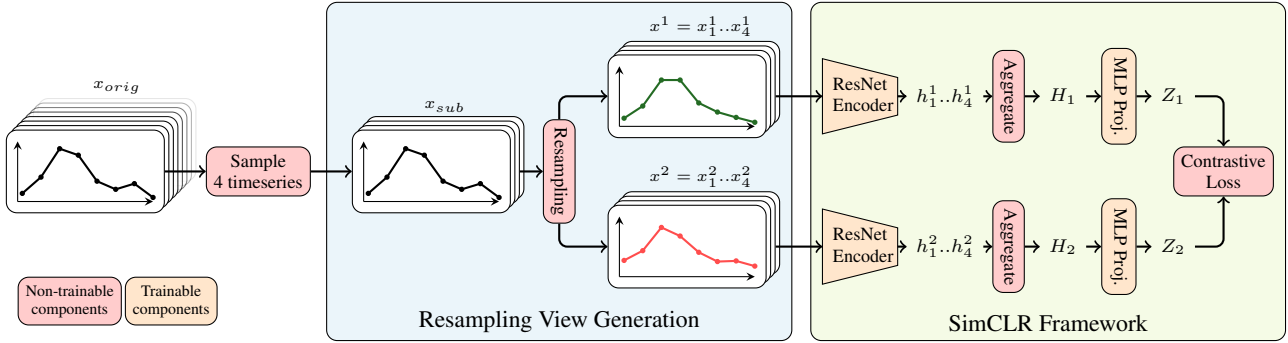
\begin{figure*}[tbp]
    \centering
    \pgfdeclarelayer{foreground}
    \pgfsetlayers{background,main,foreground}

    \begin{tikzpicture}[scale=\figscale, transform shape]
        % --- Timeseries Plot Positions ---
        \coordinate (plot_orig_pos) at (\plotOrigX, \plotOrigY); % For x_orig (original sample)
        \coordinate (plot_sub_pos) at (\plotSubX, \plotSubY);  % For x_sub (subsampled to G=4)
        \coordinate (plot1_pos) at (\plotOneX, \plotOneY);      % For x^1 (view 1 of x_sub)
        \coordinate (plot2_pos) at (\plotTwoX, \plotTwoY);     % For x^2 (view 2 of x_sub)

        % --- Original Sample x_orig Time Series Plot ---
        \begin{axis}[font=\FontSizeTiny, name=axis_orig, at={(plot_orig_pos)}, anchor=center,
            width=\axisWidth, height=\axisHeight, scale only axis,
            axis lines=middle, xlabel={ }, xmin=-0.25, xmax=7, ymin=0, ymax=37,
            xtick=\empty, ytick=\empty,
            tick label style={font=\FontSizeTiny}, xlabel style={font=\FontSizeTiny}, ylabel style={font=\FontSizeTiny},
            axis line style={draw=black},
            enlarge x limits=false, axis y line*=left,
            % title style removed as title is now a separate node
            % title removed as it's now a separate node
            axis background/.style={fill=white, rounded corners=3pt}
        ]
            \addplot[black, thick, mark=*, mark size=\plotMarkSize] coordinates { (0,5) (1,15) (2,32.5) (3,28.5) (4,12.5) (5,7.5) (6,11) (7,2.5) }; % Placeholder data for x_orig
        \end{axis}

        % --- x_orig Label (moved from axis title) ---
        % Positioned above the axis_orig node, to be above the backgrounds fitting it.
        % Uses properties from the original title style for consistency.
        
        % --- Backgrounds for x_orig ---
        \begin{scope}[on background layer]
            \node[tsBackground, opacity=0.1, fit=(axis_orig), shift={(\tsBgShiftLarge,\tsBgShiftLarge)}] (ts_orig_bg_fade_4){};
            \node[tsBackground, opacity=0.3, fit=(axis_orig), shift={(\tsBgShiftMedLarge,\tsBgShiftMedLarge)}] (ts_orig_bg_fade_3){};
            \node[tsBackground, opacity=0.5, fit=(axis_orig), shift={(\tsBgShiftMedium,\tsBgShiftMedium)}] (ts_orig_bg_fade_2){};
            \node[tsBackground, opacity=0.7, fit=(axis_orig), shift={(\tsBgShiftSmall,\tsBgShiftSmall)}] (ts_orig_bg_fade_1){};
            \node[tsBackground, opacity=0.9, fit=(axis_orig), shift={(\tsBgShiftTiny,\tsBgShiftTiny)}] (ts_orig_bg_4){};
            \node[tsBackground, fit=(axis_orig), shift={(\tsBgShiftMicro,\tsBgShiftMicro)}] (ts_orig_bg_3){};
            \node[tsBackground, fit=(axis_orig), shift={(\tsBgShiftNano,\tsBgShiftNano)}] (ts_orig_bg_2){};
            \node[tsBackground, fit=(axis_orig)] (ts_orig_bg_1){};
        \end{scope}
        \node[font=\FontSizeTiny, anchor=south, above=1pt of ts_orig_bg_fade_4.north, yshift=\axisTitleYShift] {$x_{orig}$};

        % --- Subsampled x_sub Time Series Plot (formerly x) ---
        \node[draw, rectangle, rounded corners, fill=red!20, inner sep=\componentInnerSep, minimum size=\resampleNodeMinSize, font=\FontSizeSmall] (resample_node) at ($(plot_sub_pos) + (\resampleNodeOffsetX,0)$) {\rotatebox{-90}{Resampling}};
        \begin{axis}[font=\FontSizeTiny, name=axis_sub, at={(plot_sub_pos)}, anchor=center, % Renamed from axis to axis_sub
            width=\axisWidth, height=\axisHeight, scale only axis,
            axis lines=middle, xlabel={ }, xmin=-0.25, xmax=7, ymin=0, ymax=37,
            xtick=\empty, ytick=\empty,
            tick label style={font=\FontSizeTiny}, xlabel style={font=\FontSizeTiny}, ylabel style={font=\FontSizeTiny},
            axis line style={draw=black},
            enlarge x limits=false, axis y line*=left,
            axis background/.style={fill=white, rounded corners=3pt}
        ]
            \addplot[black, thick, mark=*, mark size=\plotMarkSize] coordinates { (0,5) (1,15) (2,32.5) (3,28.5) (4,12.5) (5,7.5) (6,11) (7,2.5) };
        \end{axis}

        % --- x^1 Time Series Plot ---
        \begin{axis}[font=\FontSizeTiny, name=axis1, at={(plot1_pos)}, anchor=center,
            width=\axisWidth, height=\axisHeight, scale only axis,
            axis lines=middle, xlabel={ }, xmin=-0.25, xmax=7, ymin=0, ymax=37,
            xtick=\empty, ytick=\empty,
            tick label style={font=\FontSizeTiny}, xlabel style={font=\FontSizeTiny}, ylabel style={font=\FontSizeTiny},
            axis line style={draw=black},
            enlarge x limits=false, axis y line*=left,
            axis background/.style={fill=white, rounded corners=3pt}
        ]
            \addplot[black!30!green, thick, mark=*, mark size=\plotMarkSize] coordinates { (0,5) (1,12.5) (2,28.5) (3,28.5) (4,14.4) (5,8.8) (6,5.5) (7,2.5) };
        \end{axis}

        % --- x^2 Time Series Plot ---
        \begin{axis}[font=\FontSizeTiny, name=axis2, at={(plot2_pos)}, anchor=center,
            width=\axisWidth, height=\axisHeight, scale only axis,
            axis lines=middle, xlabel={ }, xmin=-0.25, xmax=7, ymin=0, ymax=37,
            xtick=\empty, ytick=\empty,
            tick label style={font=\FontSizeTiny}, xlabel style={font=\FontSizeTiny}, ylabel style={font=\FontSizeTiny},
            axis line style={draw=black},
            enlarge x limits=false, axis y line*=left,
            axis background/.style={fill=white, rounded corners=3pt}
        ]
            \addplot[white!30!red, thick, mark=*, mark size=\plotMarkSize] coordinates { (0,10) (1,16) (2,30.3) (3,25) (4,15) (5,9.3) (6,9.7) (7,6.5) };
        \end{axis}

        % --- Background Highlights and Connections ---

        % Encapsulate Resampling View Generation
        \begin{scope}[on background layer]
            % Top Resampling View Generation
            \node[
                draw, fill=blue!10, rounded corners, inner sep=\viewGenBgInnerSep, minimum height=\viewGenBgMinHeight,
                fit=(axis_sub) (resample_node) (axis1) (axis2), % Fit from axis_sub now
                label={[anchor=south, font=\FontSizeNormal]south: Resampling View Generation}
            ] (view_generation_bg) {};
        \end{scope}

        \begin{scope}[on background layer]
            % Backgrounds for x group
            \node[tsBackground, fit=(axis_sub), shift={(\tsBgShiftTiny,\tsBgShiftTiny)}] (ts_bg_4){}; 
            \node[tsBackground, fit=(axis_sub), shift={(\tsBgShiftMicro,\tsBgShiftMicro)}] (ts_bg_3){}; 
            \node[tsBackground, fit=(axis_sub), shift={(\tsBgShiftNano,\tsBgShiftNano)}] (ts_bg_2){}; 
            \node[tsBackground, fit=(axis_sub)] (ts_bg_1){};                         
            \node[tsBackground, fit=(axis1), shift={(\tsBgShiftTiny,\tsBgShiftTiny)}] (ts1_bg_4){};
            \node[tsBackground, fit=(axis1), shift={(\tsBgShiftMicro,\tsBgShiftMicro)}] (ts1_bg_3){};
            \node[tsBackground, fit=(axis1), shift={(\tsBgShiftNano,\tsBgShiftNano)}] (ts1_bg_2){};
            \node[tsBackground, fit=(axis1)] (ts1_bg_1){};                        
            \node[tsBackground, fit=(axis2), shift={(\tsBgShiftTiny,\tsBgShiftTiny)}] (ts2_bg_4){};
            \node[tsBackground, fit=(axis2), shift={(\tsBgShiftMicro,\tsBgShiftMicro)}] (ts2_bg_3){};
            \node[tsBackground, fit=(axis2), shift={(\tsBgShiftNano,\tsBgShiftNano)}] (ts2_bg_2){};
            \node[tsBackground, fit=(axis2)] (ts2_bg_1){};                        
        \end{scope}
        \node[font=\FontSizeTiny, anchor=south, above=1pt of ts_bg_4.north, yshift=\axisTitleYShift] {$x_{sub}$};
        \node[font=\FontSizeTiny, anchor=south, above=1pt of ts1_bg_4.north, yshift=\axisTitleYShift] {$x^1 = x^1_1 .. x^1_4$};
        \node[font=\FontSizeTiny, anchor=south, above=1pt of ts2_bg_4.north, yshift=\axisTitleYShift] {$x^2 = x^2_1 .. x^2_4$};

        % Connections for the initial subsampling step
        \path (ts_orig_bg_1.east) -- (ts_bg_1.west) node[midway, draw, rectangle, rounded corners, fill=red!20, font=\FontSizeSmall, align=center, inner sep=\componentInnerSep] (sample_node) {Sample\\4 timeseries};
        \draw[->, thick] (ts_orig_bg_1.east) -- (sample_node.west);
        \draw[->, thick] (sample_node.east) -- (ts_bg_1.west);
        % --- MLP and Loss Components (SimCLR) ---
        % Define styles for MLP/Loss components
        \tikzstyle{projector} = [draw, rectangle, rounded corners, minimum width=\projectorMinWidth, minimum height=\projectorMinHeight, fill=orange!20, align=center]
        \tikzstyle{loss} = [draw, circle, minimum size=\resampleNodeMinSize, fill=red!20, align=center]
        \tikzstyle{aggregator} = [draw, rectangle, rounded corners, fill=red!20, minimum width=\aggregatorMinWidth, inner sep=\componentInnerSep]
        \tikzstyle{componentstyle} = [draw, rectangle, rounded corners, fill=orange!10, inner sep=\componentStyleInnerSep]

        % --- SimCLR Components Scope (for x pathway) ---
        \begin{scope}[shift={($(view_generation_bg.east)+(\simclrScopeShiftX,0)$)}] 
            % --- Branch for x_1 (Top) ---
            \pic (enc1) at (0,\encOneY) {encoderPic}; 
            \node[font=\FontSizeSmall, align=center, text width=\encLabelTextWidth, below=\encLabelOffsetY of enc1-center] {ResNet\\Encoder};
            \node[font=\FontSizeSmall, right=\hSetLabelOffsetX of enc1-right, align=center] (h1_set_label) {$h^1_1 .. h^1_4$};            
            \draw[->, thick] (enc1-right) -- (h1_set_label.west);           
            \node[font=\FontSizeSmall, aggregator, right=\aggOffsetX of h1_set_label] (agg1) {\rotatebox{-90}{Aggregate}};          
            \draw[->, thick] (h1_set_label.east) -- (agg1.west);        
            \node[font=\FontSizeSmall, right=\HLabelOffsetX of agg1] (H1_label_node) {$H_1$}; % Renamed H1_label to H1_label_node to avoid conflict if H1_label was a style
            \draw[->, thick] (agg1.east) -- (H1_label_node.west);
            \node[font=\FontSizeSmall, projector, right=\projOffsetX of H1_label_node] (proj1) {\rotatebox{-90}{MLP Proj.}};
            \node[font=\FontSizeSmall, right=\ZLabelOffsetX of proj1] (Z1_label) {$Z_1$};
            \coordinate (Z1_fit_extend) at ($(Z1_label.east) + (\ZLabelFitExtend,0)$); 
            \draw[->, thick] (H1_label_node.east) -- (proj1.west); % Re-added arrow H -> MLP

            % --- Branch for x_2 (Bottom) ---
            \pic (enc2) at (0,\encTwoY) {encoderPic};    % Y adjusted for x2
            % \node[font=\FontSizeSmall, align=center, text width=\encLabelTextWidth, below=\encLabelOffsetY of enc2-center] {ResNet Encoder};
            \node[font=\FontSizeSmall, align=center, text width=\encLabelTextWidth, below=\encLabelOffsetY of enc2-center] {ResNet\\Encoder};
            \node[font=\FontSizeSmall, right=\hSetLabelOffsetX of enc2-right, align=center] (h2_set_label) {$h^2_1 .. h^2_4$};            
            \draw[->, thick] (enc2-right) -- (h2_set_label.west);
            \node[font=\FontSizeSmall, aggregator, right=\aggOffsetX of h2_set_label] (agg2) {\rotatebox{-90}{Aggregate}};          
            \draw[->, thick] (h2_set_label.east) -- (agg2.west);
            \node[font=\FontSizeSmall, right=\HLabelOffsetX of agg2] (H2_label_node) {$H_2$}; % Renamed H2_label to H2_label_node
            \draw[->, thick] (agg2.east) -- (H2_label_node.west);
            \node[font=\FontSizeSmall, projector, right=\projOffsetX of H2_label_node] (proj2) {\rotatebox{-90}{MLP Proj.}};
            \node[font=\FontSizeSmall, right=\ZLabelOffsetX of proj2] (Z2_label) {$Z_2$};
            \coordinate (Z2_fit_extend) at ($(Z2_label.east) + (\ZLabelFitExtend,0)$); 
            \draw[->, thick] (H2_label_node.east) -- (proj2.west); % Re-added arrow H -> MLP

            % Connections for SimCLR scope (x pathway)
            \draw[->, thick] ($(ts_bg_1.east) + (\arrowToResampleNodeOffsetX,0)$) -- (resample_node.west);
            \draw[->, thick, rounded corners=3pt] (resample_node) |- (ts1_bg_1);
            \draw[->, thick, rounded corners=3pt] (resample_node) |- (ts2_bg_1);
            \draw[->, thick] ($(ts1_bg_1.east) + (\arrowToEncoderOffsetX,0)$) -- (enc1-left);
            \draw[->, thick] ($(ts2_bg_1.east) + (\arrowToEncoderOffsetX,0)$) -- (enc2-left);

            % --- Contrastive Loss ---
            \node[font=\FontSizeSmall, draw, rectangle, rounded corners, fill=red!20, align=center, inner sep=\componentInnerSep] (contrastive_loss) at ($(Z1_label.east)!0.5!(Z2_label.east) + (\contrastiveLossOffsetX,0)$) {Contrastive\\Loss};
            \draw[->, thick, rounded corners=3pt] (Z1_label.east) -| (contrastive_loss);
            \draw[->, thick, rounded corners=3pt] (Z2_label.east) -| (contrastive_loss);
            \draw[->, thick] (proj2.east) -- (Z2_label.west); % Added arrow
            \draw[->, thick] (proj1.east) -- (Z1_label.west); % Added arrow

            % Background for SimCLR Framework
            \begin{scope}[on background layer]
                \node[draw, fill=my_green!10, rounded corners, inner sep=\simclrBgInnerSep, minimum height=\simclrBgMinHeight,
                      fit=(enc1-left)(h1_set_label)(agg1)(H1_label_node)(proj1)(Z1_label)(Z1_fit_extend) % Top branch, used H1_label_node
                          (enc2-left)(h2_set_label)(agg2)(H2_label_node)(proj2)(Z2_label)(Z2_fit_extend) % Bottom branch, used H2_label_node
                          (contrastive_loss.east), % Ensure fit includes the new node
                      label={[anchor=south, font=\FontSizeNormal]south:SimCLR Framework}
                ] (simclr_bg_scoped) {};
            \end{scope}
        \end{scope}

        % --- Legend (New Style: Inside ssl_pretraining_bg box) ---
        \node (legend_nontrainable) [font=\FontSizeTiny, rectangle, draw, fill=red!20, rounded corners, align=center, inner sep=\componentInnerSep, anchor=north west, at={($(axis_orig.south west) + (0,\legendNontrainableOffsetY)$)}] {Non-trainable\\components};
        \node (legend_trainable) [font=\FontSizeTiny, rectangle, draw, fill=orange!20, rounded corners, align=center, inner sep=\componentInnerSep, right=\legendTrainableSpacing of legend_nontrainable, anchor=west] {Trainable\\components};

        % Encapsulate full SSL pretraining
        % \begin{scope}[on background layer]
        %     \node[
        %         draw, rounded corners, inner sep=\sslBgInnerSep,
        %         fit=(view_generation_bg) (simclr_bg_scoped) (legend_nontrainable) (legend_trainable) (ts_orig_bg_1), % Removed subsampling_bg from main fit
        %         label={[anchor=south, font=\FontSizeNormal, yshift=\sslLabelYShift]south:Self supervised contrastive pretraining}
        %     ] (ssl_pretraining_bg) {};
        % \end{scope}
    \end{tikzpicture}
    \caption{Self-Supervised Contrastive retraining. For each sample $x_{orig}$ in a batch, 4 time series are randomly sampled from it into $x_{sub}$. The resampling augmentation is then applied to create two views, $x^1$ and $x^2$. Each of the 4 time series per view per sample is individually fed to a ResNet encoder, producing 4 embeddings per view per sample into a representation space $h$ that will be used for downstream tasks. For each view, these 4 embeddings are then aggregated by averaging into a single representation ($H_1$ and $H_2$). These aggregated representations are further projected by an MLP to $Z_1$ and $Z_2$, on which the contrastive loss is computed with other projections from other samples in the batch. The diagram illustrates the SimCLR framework but any contrastive learning framework that rely on positive pairs can be used.}
    \label{fig:resampling_simclr_framework}
\end{figure*}

Figure \ref{fig:resampling_simclr_framework} describes our architecture. Following a standard contrastive learning architecture \cite{chen2021exploring}, our model consists of an encoder network followed by a projection head. The encoder maps the input time series to a representation space we use for downstream tasks, while the projection head further transforms these representations for optimization of the contrastive loss. In Section \ref{sec:contrastive_learning_framework_comparison} we compare performance on SimCLR \cite{chen2020simple}, MoCo \cite{he2019momentum}, BYOL \cite{grill2020bootstrap} and VICReg \cite{bardes2021vicreg} frameworks. We refer readers to SimSiam \cite{chen2021exploring} for diagrams comparing different contrastive learning architectures.

We use a ResNet encoder adapted for time series \cite{wang2016time}, configured with 256 filters in the first convolutional layer. The encoder outputs 512-dimensional embeddings. Considering that for all datasets a sample consists of a set of multiple time series, we reuse the approach from \cite{saget2024learning} to aggregate multiple time series embeddings from the same sample into one: during the forward pass, $G$ (for group) pixel time series (typically $G=4$) are randomly selected per sample during training. Each of the $G$ series is processed independently by the shared ResNet blocks and global pooling, yielding $G$ embeddings of dimension 512 per sample. An extra adaptive average pooling layer (appended after the ResNet encoder output) aggregates them into a single 512-dimensional vector per sample. This aggregation layer fuses multiple time series embeddings into one by averaging along the group dimension. Empirically, setting $G=4$ captures intra-sample time series variability without over-smoothing discriminative features (see \cite{saget2024learning}). Note that for datasets without multiple time series per sample (when $G=1$), the aggregation layer is equivalent to the identity operation and can be discarded.

We use a 2-layer Multi-Layer Perceptron (MLP) projection head with a hidden dimension of 512 and an output dimension of 128.

\subsection{Training and evaluation protocol}

\textbf{Pretraining:} Models are trained using Stochastic Gradient Descent (SGD) with 5e-4 weight decay and 0.9 momentum. For frameworks using momentum encoders (MoCo and BYOL), the target model weights are updated with a momentum of 0.996. We employ a one-cycle learning rate policy starting at 2e-3, increasing to 5e-2 for the first 20\% of training, then decreasing to 5e-5. Training runs for 50k steps with a batch size of 1024. The best checkpoint is selected based on performance on a small validation dataset, evaluated every 1000 steps.

\textbf{Downstream Evaluation:} We evaluate the pretrained encoder in two ways:
\begin{itemize}
    \item \textbf{Linear Evaluation:} A logistic regression (max\_iter: 2000, tol: 1e-5, C: 1.0) is trained on frozen encoder features.
    \item \textbf{Finetuning:} A 2-layer MLP with a hidden dimension of 256, ReLU activation, and 20\% dropout is added to the encoder. The MLP is trained alone for 10 epochs (encoder frozen), then the full model is finetuned for 100 epochs. The MLP learning rate is 1e-3 and the encoder learning rate is 2e-5 with 5e-4 weight decay.
\end{itemize}

We train and evaluate on a single NVIDIA RTX 4090 GPU. Pretraining takes approximately 6 hours, finetuning takes 2-3 minutes, and linear evaluation takes a few seconds.

As datasets include multiple time series per sample, the final prediction of a sample is obtained through majority voting over individual time series predictions.

\section{Results}
\label{sec:results}

\subsection{Contrastive learning framework comparison}
\label{sec:contrastive_learning_framework_comparison}

Table \ref{tab:contrastive_learning_framework_comparison} shows our experimental results comparing SimCLR, MoCo, BYOL, and VICReg contrastive learning frameworks on the FranceCrops dataset. VICReg achieved the best results with 72\% accuracy when using resampling augmentation. Our resampling augmentation outperforms other augmentations across all contrastive learning frameworks tested.

However, we observed important differences in training stability between frameworks. Notably, representation collapse occurs more frequently with time series data compared to images. We hypothesize that this is due to the lower dimensionality of time series data, making it easier for models to find trivial solutions that map all inputs to the same representation. While VICReg achieves the highest accuracy, we found it occasionally diverges on other datasets and requires careful hyperparameter tuning. In contrast, SimCLR shows more stable training behavior despite achieving slightly lower accuracy (69\%). For this reason, we use SimCLR for the remainder of our experiments (except for FranceCrops where VICReg is stable), prioritizing reliable training over marginal performance gains.

\begin{table}[hbtp]
\caption{Mean Test Accuracy (\%) of Logistic Regressions trained on features extracted by encoders trained with various contrastive learning frameworks. Results averaged over 20 runs with different training sets on the FranceCrops Dataset with 10 labeled samples per class. All standard deviations are $\leq 1$. A logistic regression on raw data achieve 52\% accuracy.}
\begin{center}
\begin{tabular}{c||c|c|c|c}
    & \multicolumn{4}{c}{\textbf{SSL Framework}} \\
    \cline{2-5} 
    \textbf{Aug.} & \textbf{\textit{SimCLR}}& \textbf{\textit{BYOL}}& \textbf{\textit{VICReg}}& \textbf{\textit{MoCo}} \\
    \hline
    \hline
    Jittering & 49 & 48 & 53 & 48  \\
    Resizing & 54 & 56 & 65 & 57  \\
    Masking & 61 & 62 & 67 & 63  \\
    Resampling & \textbf{69} & \textbf{69} & \textbf{72} & \textbf{68}  \\
\end{tabular}
\label{tab:contrastive_learning_framework_comparison}
\end{center}
\end{table}
\subsection{Label Efficiency}
\label{sec:label_efficiency}

To evaluate how well our approach performs with limited labeled data, we conducted experiments across three datasets with varying numbers of labeled samples per class, from very few (5) to relatively many (100). Results in Table \ref{tab:label_efficiency_all_datasets} show that our resampling augmentation consistently outperforms all other approaches across all sample sizes and datasets. The improvement is most significant with few labels (5-20 samples), achieving up to 23 percentage points over raw features on FranceCrops (44\% to 67\% with 5 samples). Importantly, this performance advantage is maintained even as more labeled data becomes available, showing that our method can effectively leverage additional supervision without compromising its representational power.

In contrast, simpler augmentations like jittering show limited scalability. While beneficial with few labels (5-10 samples), it performs worse than raw features with more labels, suggesting it oversimplifies representations, making them unable to capture the full complexity of the data when more supervision is available.

Time masking performs second-best. Like our resampling approach, it creates views by removing information from the original time series. This connection is particularly interesting as it bridges the gap between contrastive learning and masking methods: both rely on learning from incomplete temporal information, but while masking methods reconstruct the missing values, our approach learns to match representations of different incomplete views. 

Despite many similarities with FranceCrops, we observe lower performance on PASTIS, likely due to differences in cloud filtering, a more challenging set of crop types, and a smaller pretraining dataset.

\begin{table}[htb]
\caption{Mean Test Accuracy (\%) of Logistic Regressions trained on features extracted by a SimCLR encoder (VICReg for FranceCrops). Results averaged over 20 runs with different training sets. All standard deviations are $\leq 1$ unless specified in parentheses.}

\begin{center}
\begin{subtable}{\columnwidth}
\caption{FranceCrops}
\begin{center}
\begin{tabular}{c||c|c|c|c|c}
\textbf{Augmentation} & \multicolumn{5}{c}{\textbf{Samples per Class (Training)}} \\
\cline{2-6}
\textbf{Strategy} & \textbf{\textit{5}}& \textbf{\textit{10}}& \textbf{\textit{20}}& \textbf{\textit{50}}& \textbf{\textit{100}} \\
\hline
\hline
Raw Features & 44 & 52 & 61 & 70 & 76 \\
\hline
Jittering & 49 & 53 & 59 & 65 & 69 \\
Resizing & 60 & 65 & 69 & 74 & 77 \\
Masking & 62 & 67 & 73 & 78 & 81 \\
Resampling & \textbf{67} & \textbf{72} & \textbf{76} & \textbf{80} & \textbf{83} \\
\end{tabular}
\end{center}
\end{subtable}

\vspace{3mm}
\begin{subtable}{\columnwidth}
\caption{FranceCrops Centre-val de Loire}
\begin{center}
\begin{tabular}{c||c|c|c|c|c}
\textbf{Augmentation} & \multicolumn{5}{c}{\textbf{Samples per Class (Training)}} \\
\cline{2-6}
\textbf{Strategy} & \textbf{\textit{5}}& \textbf{\textit{10}}& \textbf{\textit{20}}& \textbf{\textit{50}}& \textbf{\textit{100}} \\
\hline
\hline
Raw Features & 49 (2) & 56 (2) & 64 & 74 & 79 \\
\hline
Jittering & 52 & 58 & 63 & 68 & 71 \\
Resizing & 59 & 63 & 68 & 73 & 76 \\
Masking & 62 & 68 & 72 & 77 & 80 \\
Resampling & \textbf{65} & \textbf{71} & \textbf{75} & \textbf{79} & \textbf{82} \\
\end{tabular}
\end{center}
\end{subtable}

\vspace{3mm}
\begin{subtable}{\columnwidth}
\caption{Pastis}
\begin{center}
\begin{tabular}{c||c|c|c|c|c}
\textbf{Augmentation} & \multicolumn{5}{c}{\textbf{Samples per Class (Training)}} \\
\cline{2-6}
\textbf{Strategy} & \textbf{\textit{5}}& \textbf{\textit{10}}& \textbf{\textit{20}}& \textbf{\textit{50}}& \textbf{\textit{100}} \\
\hline
\hline
Raw Features & 24 & 28 & 32 & 37 & 40 \\
\hline
Jittering & 23 & 26 & 29 & 32 & 33 \\
Resizing & 26 & 29 & 33 & 36 & 37 \\
Masking & 37 & 41 & 45 & 47 & 48 \\
Resampling & \textbf{38} & \textbf{42} & \textbf{46} & \textbf{49} & \textbf{50} \\
\end{tabular}
\end{center}
\end{subtable}
\label{tab:label_efficiency_all_datasets}
\end{center}
\end{table}

\subsection{Image Time Series Task - S2-Agri100 Results}

We evaluate our model's ability to generalize across different geographical regions by pretraining on SITS-Former's California data and testing on French agricultural parcels from the S2-Agri100 dataset. Our approach does not leverage any spatial information beyond the assumption that pixels within the same patch share the same label \cite{saget2024learning}. Following \citeauthor{tseng2023lightweight} \yrcite{tseng2023lightweight}, we use 100 parcels per class for training, a validation set for early stopping, and the remaining parcels for testing.

Table \ref{tab:s2agri_results_pretrain} shows the corresponding performance after finetuning models that were first pretrained in a self-supervised manner on the SITS-Former dataset. Despite our model's simpler architecture, which ignores spatial information and temporal positions, it achieves superior performance. Following \citeauthor{tseng2023lightweight} \yrcite{tseng2023lightweight}, we also report in Table \ref{tab:s2agri_results_no_pretrain} results when training each model from random initialization (without pretraining) using only the S2-Agri100 training set. 

\begin{table}[htbp]
\caption{Results on the S2-Agri100 dataset after self-supervised pretraining on SITS-Former and finetuning on S2-Agri100. We report Overall Accuracy (OA), Kappa Cohen score (K) and macro-F1 score following \citeauthor{tseng2023lightweight} \yrcite{tseng2023lightweight}. Averages over three runs.}
\begin{center}
\begin{tabular}{c||c|c|c|c}
\textbf{Model}    & \textbf{M Params} & \textbf{OA}   & \textbf{K}    & \textbf{F1}   \\
\hline\hline
SITS-Former       & 2.5               & 67.03         & 0.56          & 42.83         \\
Presto            & 0.4               & 68.89         & 0.58          & 40.41         \\
Ours              & 8.2               & \textbf{70.15}& \textbf{0.60}& \textbf{44.14}\\
\end{tabular}
\label{tab:s2agri_results_pretrain}
\end{center}
\end{table}

\begin{table}[htbp]
\caption{Results on the S2-Agri100 dataset when training from random initialization (no pretraining).}
\begin{center}
\begin{tabular}{c||c|c|c|c}
\textbf{Model}    & \textbf{M Params} & \textbf{OA}   & \textbf{K}    & \textbf{F1}   \\
\hline\hline
SITS-Former       & 2.5               & \textbf{65.13}         & \textbf{0.55}          & \textbf{42.12}         \\
Presto            & 0.4               & 45.98         & 0.35          & 27.45         \\
Ours              & 8.2               & 62.40         & 0.52          & 40.21         \\
\end{tabular}
\label{tab:s2agri_results_no_pretrain}
\end{center}
\end{table}

\subsection{Impact of Pretraining Data Distribution}
\label{sec:pretraining_distribution}

\begin{table*}[h!tbp]
\caption{Results when unsupervised pretraining on S2-Agri100 versus pretraining on SITS-Former illustrate the importance of learning from data with similar distribution as the target task. Average over three runs.}
\begin{center}
\begin{tabular}{c||c|c|c|c|c}
\textbf{Model} & \textbf{Pretrain Data} & \textbf{Eval. Type} & \textbf{OA} & \textbf{K} & \textbf{F1} \\
\hline\hline
SITS-Former & \multirow{3}{*}{SITS-Former} & Finetuning                  & 67.03         & 0.56          & 42.83         \\
Presto      &                              & Finetuning                  & 68.89         & 0.58          & 40.41         \\
Ours        &                              & Finetuning                  & 70.15         & 0.60          & 44.14         \\
\hline
Ours        & \multirow{2}{*}{S2-Agri100}  & Finetuning                  & \textbf{76.84}& \textbf{0.68}& \textbf{48.55}\\
Ours        &                              & Linear (log. reg.)          & 74.30         & 0.65          & 48.36         \\
\end{tabular}
\label{tab:feature_learning}
\end{center}
\end{table*}

In this experiment, we investigate the impact of pretraining data distribution. We compare pretraining on data from the same distribution as the target task versus pretraining on data from a different distribution.  While we used the SITS-Former dataset (California) for pretraining and S2-Agri100 (France) for evaluation in our previous experiment, we both pretrain and evaluate on S2-Agri100 in this experiment.

% While this setup is not a fair comparison to other methods since we use test data during pretraining, it provides crucial insights about the importance of distribution matching between pretraining and target task data.

The S2-Agri100 train split is too small (1500 samples) for pretraining, so we pretrain on 70\% of the large S2-Agri100 test split in an unsupervised manner and evaluate on the remaining 30\% (over 40,000 samples). We still use the standard S2-Agri100 training split for finetuning and linear evaluation.

Table \ref{tab:feature_learning} shows that both approaches significantly outperform all previous methods, including our model pretrained on SITS-Former. This improvement should not be attributed to our specific model architecture - similar gains might be observed if other models like Presto were pretrained in the same way. Rather, these results illustrate two general principles: first, the importance of pretraining on data from the same distribution as the target task, and second, as shown by the minimal gap between logistic regression (74.30\% OA) and full finetuning (76.84\% OA), most of the performance comes from the quality of the learned features rather than the complexity of the supervised classifier.

These results have important implications for practical applications. First, they highlight that matching the distribution between pretraining data and target task is crucial for optimal performance. Second, they suggest that feature learning can be effectively decoupled from classification: strong features can be learned from unlabeled data, while a simple classifier trained on these features with limited labeled data can achieve excellent performance. In other words, collecting large quantities of unlabeled data from the target domain can be as valuable as obtaining small quantities of labels.

\section{Conclusion}
\label{sec:conclusion}

Our approach significantly reduces the need for labeled data across all tested datasets and outperforms other traditional augmentations like jittering, masking, and resizing. With just 5 labeled samples per class, our method achieves performance comparable to training the same model on raw features with 20-50 samples per class, representing a 4-10x reduction in required labeled data.

The effectiveness of our resampling augmentation stems from its ability to create meaningful positive pairs while preserving temporal structure. Notably, it requires only two hyperparameters ($T_{up}$ and $T_{sub}$) that we set to natural values ($T_{up} = 2 \times T$ and $T_{sub} = T / 2$) and did not optimize.

However, our approach has limitations. It requires time series with a high temporal sampling rate relative to the frequency of meaningful events. This assumption holds well for remote sensing data, where Sentinel-2's 5-day revisit time captures most agricultural and land cover changes that typically occur over weeks or months. For datasets with rare or high-frequency events, the subsampling might lose critical information.

Our experiments on S2-Agri100 showed that features learned from unlabeled data can be more important for performance than an advanced classifier. The minimal gap between logistic regression and full finetuning performance suggests that when domain-specific unlabeled data is available, strong results can be achieved with simple linear classifiers on pretrained features.

This study suggests several directions for future work:
\begin{itemize}
    \item Evaluating the resampling augmentation in standard supervised learning settings, beyond its current use in contrastive learning.
    \item Exploring the applicability of our augmentation to sequential data from other domains beyond remote sensing, with comparable and different temporal patterns.
    \item Extending our method to a Remote Sensing Foundation Model framework supporting variable-length sequences, non-uniform sampling, multiple modalities, Earth-wide pretraining, and diverse downstream applications beyond crop classification.
    \item Incorporating spatial context instead of purely temporal analysis of individual pixels or pixel-sets.
\end{itemize}

\section*{Impact Statement}

This paper presents work whose goal is to advance the field of Machine Learning by improving data augmentation strategies for self-supervised contrastive learning on remote sensing time series. Our resampling augmentation contributes towards foundation models that can leverage vast amounts of unlabeled satellite image time series while significantly reducing the need for labeled data. By enhancing label efficiency, this approach can lower the barrier to entry for resource-constrained practitioners and promote broader use of Earth observation for environmental monitoring, agriculture, and natural disaster response.

\section*{Acknowledgment}

We thank the French National Research Agency (ANR) for the Ph.D. funding through the ANR ArtIC project and supporting this research as part of the ANR HERELLES project. We also express our gratitude to the European Space Agency (ESA) and the Copernicus programme for making Sentinel-2 data freely accessible to the scientific community. We acknowledge the use of Claude 3.7/4 Sonnet (Anthropic), GPT-4o/o4-mini/o3 (OpenAI) and Gemini-2.5-Pro (Google) large language models to assist writing the code used to conduct the experiments, and editing this article for grammar checking, polishing, and formatting. 

\bibliographystyle{icml2025}
\bibliography{ijcnn}

\begin{thebibliography}{27}
\providecommand{\natexlab}[1]{#1}
\providecommand{\url}[1]{\texttt{#1}}
\expandafter\ifx\csname urlstyle\endcsname\relax
  \providecommand{\doi}[1]{doi: #1}\else
  \providecommand{\doi}{doi: \begingroup \urlstyle{rm}\Url}\fi

\bibitem[Bardes et~al.(2021)Bardes, Ponce, and LeCun]{bardes2021vicreg}
Bardes, A., Ponce, J., and LeCun, Y.
\newblock Vicreg: Variance-invariance-covariance regularization for self-supervised learning.
\newblock \emph{arXiv preprint arXiv:2105.04906}, 2021.

\bibitem[Caron et~al.(2020)Caron, Misra, Mairal, Goyal, Bojanowski, and Joulin]{caron2020unsupervised}
Caron, M., Misra, I., Mairal, J., Goyal, P., Bojanowski, P., and Joulin, A.
\newblock Unsupervised learning of visual features by contrasting cluster assignments.
\newblock \emph{Advances in neural information processing systems}, 33:\penalty0 9912--9924, 2020.

\bibitem[Chen et~al.(2020{\natexlab{a}})Chen, Kornblith, Norouzi, and Hinton]{chen2020simple}
Chen, T., Kornblith, S., Norouzi, M., and Hinton, G.
\newblock A simple framework for contrastive learning of visual representations.
\newblock In \emph{International conference on machine learning}, pp.\  1597--1607. PMLR, 2020{\natexlab{a}}.

\bibitem[Chen et~al.(2020{\natexlab{b}})Chen, Kornblith, Swersky, Norouzi, and Hinton]{chen2020big}
Chen, T., Kornblith, S., Swersky, K., Norouzi, M., and Hinton, G.~E.
\newblock Big self-supervised models are strong semi-supervised learners.
\newblock \emph{Advances in neural information processing systems}, 33:\penalty0 22243--22255, 2020{\natexlab{b}}.

\bibitem[Chen \& He(2021)Chen and He]{chen2021exploring}
Chen, X. and He, K.
\newblock Exploring simple siamese representation learning.
\newblock In \emph{Proceedings of the IEEE/CVF conference on computer vision and pattern recognition}, pp.\  15750--15758, 2021.

\bibitem[Cheng et~al.(2023)Cheng, Liu, Liu, Zhang, Zhang, and Chen]{cheng2023timemae}
Cheng, M., Liu, Q., Liu, Z., Zhang, H., Zhang, R., and Chen, E.
\newblock Timemae: Self-supervised representations of time series with decoupled masked autoencoders.
\newblock \emph{arXiv preprint arXiv:2303.00320}, 2023.

\bibitem[Cong et~al.(2022)Cong, Khanna, Meng, Liu, Rozi, He, Burke, Lobell, and Ermon]{cong2022satmae}
Cong, Y., Khanna, S., Meng, C., Liu, P., Rozi, E., He, Y., Burke, M., Lobell, D., and Ermon, S.
\newblock Satmae: Pre-training transformers for temporal and multi-spectral satellite imagery.
\newblock \emph{Advances in Neural Information Processing Systems}, 35:\penalty0 197--211, 2022.

\bibitem[Drusch et~al.(2012)Drusch, Del~Bello, Carlier, Colin, Fernandez, Gascon, Hoersch, Isola, Laberinti, Martimort, et~al.]{drusch2012sentinel}
Drusch, M., Del~Bello, U., Carlier, S., Colin, O., Fernandez, V., Gascon, F., Hoersch, B., Isola, C., Laberinti, P., Martimort, P., et~al.
\newblock Sentinel-2: Esa's optical high-resolution mission for gmes operational services.
\newblock \emph{Remote sensing of Environment}, 120:\penalty0 25--36, 2012.

\bibitem[Garnot \& Landrieu(2021)Garnot and Landrieu]{garnot2021panoptic}
Garnot, V. S.~F. and Landrieu, L.
\newblock Panoptic segmentation of satellite image time series with convolutional temporal attention networks.
\newblock In \emph{Proceedings of the IEEE/CVF International Conference on Computer Vision}, pp.\  4872--4881, 2021.

\bibitem[Garnot et~al.(2020)Garnot, Landrieu, Giordano, and Chehata]{garnot2020satellite}
Garnot, V. S.~F., Landrieu, L., Giordano, S., and Chehata, N.
\newblock Satellite image time series classification with pixel-set encoders and temporal self-attention.
\newblock In \emph{Proceedings of the IEEE/CVF Conference on Computer Vision and Pattern Recognition}, pp.\  12325--12334, 2020.

\bibitem[Gascon et~al.(2017)Gascon, Bouzinac, Th{\'e}paut, Jung, Francesconi, Louis, Lonjou, Lafrance, Massera, Gaudel-Vacaresse, et~al.]{gascon2017copernicus}
Gascon, F., Bouzinac, C., Th{\'e}paut, O., Jung, M., Francesconi, B., Louis, J., Lonjou, V., Lafrance, B., Massera, S., Gaudel-Vacaresse, A., et~al.
\newblock Copernicus sentinel-2a calibration and products validation status.
\newblock \emph{Remote Sensing}, 9\penalty0 (6):\penalty0 584, 2017.

\bibitem[Grill et~al.(2020)Grill, Strub, Altch{\'e}, Tallec, Richemond, Buchatskaya, Doersch, Avila~Pires, Guo, Gheshlaghi~Azar, et~al.]{grill2020bootstrap}
Grill, J.-B., Strub, F., Altch{\'e}, F., Tallec, C., Richemond, P., Buchatskaya, E., Doersch, C., Avila~Pires, B., Guo, Z., Gheshlaghi~Azar, M., et~al.
\newblock Bootstrap your own latent-a new approach to self-supervised learning.
\newblock \emph{Advances in neural information processing systems}, 33:\penalty0 21271--21284, 2020.

\bibitem[Guo et~al.(2024)Guo, Lao, Dang, Zhang, Yu, Ru, Zhong, Huang, Wu, Hu, et~al.]{guo2024skysense}
Guo, X., Lao, J., Dang, B., Zhang, Y., Yu, L., Ru, L., Zhong, L., Huang, Z., Wu, K., Hu, D., et~al.
\newblock Skysense: A multi-modal remote sensing foundation model towards universal interpretation for earth observation imagery.
\newblock In \emph{Proceedings of the IEEE/CVF Conference on Computer Vision and Pattern Recognition}, pp.\  27672--27683, 2024.

\bibitem[He et~al.(2019)He, Fan, Wu, Xie, and Girshick]{he2019momentum}
He, K., Fan, H., Wu, Y., Xie, S., and Girshick, R.
\newblock Momentum contrast for unsupervised visual representation learning. arxiv e-prints, art.
\newblock \emph{arXiv preprint arXiv:1911.05722}, 2019.

\bibitem[He et~al.(2022)He, Chen, Xie, Li, Doll{\'a}r, and Girshick]{he2022masked}
He, K., Chen, X., Xie, S., Li, Y., Doll{\'a}r, P., and Girshick, R.
\newblock Masked autoencoders are scalable vision learners.
\newblock In \emph{Proceedings of the IEEE/CVF conference on computer vision and pattern recognition}, pp.\  16000--16009, 2022.

\bibitem[Henaff(2020)]{henaff2020data}
Henaff, O.
\newblock Data-efficient image recognition with contrastive predictive coding.
\newblock In \emph{International conference on machine learning}, pp.\  4182--4192. PMLR, 2020.

\bibitem[Jakubik et~al.(2023)Jakubik, Roy, Phillips, Fraccaro, Godwin, Zadrozny, Szwarcman, Gomes, Nyirjesy, Edwards, et~al.]{jakubik2023foundation}
Jakubik, J., Roy, S., Phillips, C., Fraccaro, P., Godwin, D., Zadrozny, B., Szwarcman, D., Gomes, C., Nyirjesy, G., Edwards, B., et~al.
\newblock Foundation models for generalist geospatial artificial intelligence.
\newblock \emph{CoRR}, 2023.

\bibitem[Liu et~al.(2024)Liu, Alavi, Li, and Zhang]{liu2024guidelines}
Liu, Z., Alavi, A., Li, M., and Zhang, X.
\newblock Guidelines for augmentation selection in contrastive learning for time series classification.
\newblock \emph{arXiv preprint arXiv:2407.09336}, 2024.

\bibitem[Manas et~al.(2021)Manas, Lacoste, Gir{\'o}-i Nieto, Vazquez, and Rodriguez]{manas2021seasonal}
Manas, O., Lacoste, A., Gir{\'o}-i Nieto, X., Vazquez, D., and Rodriguez, P.
\newblock Seasonal contrast: Unsupervised pre-training from uncurated remote sensing data.
\newblock In \emph{Proceedings of the IEEE/CVF International Conference on Computer Vision}, pp.\  9414--9423, 2021.

\bibitem[Reed et~al.(2022)Reed, Gupta, Li, Brockman, Funk, Clipp, Candido, UyttenDAele, and Darrell]{reed2022scale}
Reed, C., Gupta, R., Li, S., Brockman, S., Funk, C., Clipp, B., Candido, S., UyttenDAele, M., and Darrell, T.
\newblock Scale-mae: A scale-aware masked autoencoder for multiscale geospatial representation learning. 2023 ieee.
\newblock In \emph{CVF International Conference on Computer Vision (ICCV)}, pp.\  4065--4076, 2022.

\bibitem[Saget et~al.(2024)Saget, Lafabregue, Cornu{\'e}jols, and Gan{\c{c}}arski]{saget2024learning}
Saget, A., Lafabregue, B., Cornu{\'e}jols, A., and Gan{\c{c}}arski, P.
\newblock Learning from few labeled time series with segment-based self-supervised learning: application to remote-sensing.
\newblock In \emph{Proceedings of SPAICE2024: The First Joint European Space Agency/IAA Conference on AI in and for Space}, pp.\  275--279, 2024.

\bibitem[Tseng et~al.(2023)Tseng, Cartuyvels, Zvonkov, Purohit, Rolnick, and Kerner]{tseng2023lightweight}
Tseng, G., Cartuyvels, R., Zvonkov, I., Purohit, M., Rolnick, D., and Kerner, H.
\newblock Lightweight, pre-trained transformers for remote sensing timeseries.
\newblock \emph{arXiv preprint arXiv:2304.14065}, 2023.

\bibitem[Wang et~al.(2022)Wang, Braham, Xiong, Liu, Albrecht, Zhu, et~al.]{wang2022ssl4eo}
Wang, Y., Braham, N., Xiong, Z., Liu, C., Albrecht, C., Zhu, X., et~al.
\newblock Ssl4eo-s12: a large-scale multi-modal, multi-temporal dataset for self-supervised learning in earth observation. arxiv.
\newblock \emph{arXiv preprint arXiv:2211.07044}, 10, 2022.

\bibitem[Wang et~al.(2023)Wang, Albrecht, Braham, Liu, Xiong, and Zhu]{wang2023decur}
Wang, Y., Albrecht, C.~M., Braham, N. A.~A., Liu, C., Xiong, Z., and Zhu, X.~X.
\newblock Decur: decoupling common \& unique representations for multimodal self-supervision.
\newblock \emph{arXiv preprint arXiv:2309.05300}, 2023.

\bibitem[Wang et~al.(2016)Wang, Yan, and Oates]{wang2016time}
Wang, Z., Yan, W., and Oates, T.
\newblock Time series classification from scratch with deep neural networks: a strong baseline. corr abs/1611.06455 (2016).
\newblock \emph{arXiv preprint arXiv:1611.06455}, 2016.

\bibitem[Yuan et~al.(2022)Yuan, Lin, Liu, Hang, and Zhou]{yuan2022sits}
Yuan, Y., Lin, L., Liu, Q., Hang, R., and Zhou, Z.-G.
\newblock Sits-former: A pre-trained spatio-spectral-temporal representation model for sentinel-2 time series classification.
\newblock \emph{International Journal of Applied Earth Observation and Geoinformation}, 106:\penalty0 102651, 2022.

\bibitem[Zbontar et~al.(2021)Zbontar, Jing, Misra, LeCun, and Deny]{zbontar2021barlow}
Zbontar, J., Jing, L., Misra, I., LeCun, Y., and Deny, S.
\newblock Barlow twins: Self-supervised learning via redundancy reduction.
\newblock In \emph{International conference on machine learning}, pp.\  12310--12320. PMLR, 2021.

\end{thebibliography}

\end{document}